\documentclass[journal]{IEEEtran}
\usepackage{cite}
\usepackage{amsmath,amssymb,amsfonts}
\usepackage{algorithmic}
\usepackage{graphicx}
\usepackage{textcomp}
\usepackage{caption2}
\usepackage{booktabs}
\usepackage{multirow}
\usepackage{xcolor,colortbl}
\definecolor{mygray}{gray}{.9}
\def\BibTeX{{\rm B\kern-.05em{\sc i\kern-.025em b}\kern-.08em
    T\kern-.1667em\lower.7ex\hbox{E}\kern-.125emX}}
\begin{document}
\title{DecGAN: Decoupling Generative Adversarial Network detecting abnormal neural circuits for Alzheimer's disease}
\author{Junren Pan, Baiying Lei, Shuqiang Wang, Bingchuan Wang, Yong Liu, Yanyan Shen
\thanks{Junren Pan, Yanyan Shen and Shuqiang Wang are with the Shenzhen Institutes of Advanced Technology, Chinese Academy of Sciences, Shenzhen 518055, China.Email:sq.wang@siat.ac.cn}
\thanks{Baiying Lei is with the School of Biomedical Engineering, Shenzhen University, Shenzhen 518060, China.}
\thanks{Yong Liu is with the Gaoling School of Artificial Intelligence, Renmin University of China, Beijing, 100000, China.}
\thanks{Bingchuan Wang is with the School of Automation,Central South University, Changsha 410083, China}}
\maketitle

\begin{abstract}
One of the main reasons for Alzheimer's disease (AD) is the disorder of some neural circuits.
Existing methods for AD prediction have achieved great success,
however, detecting abnormal neural circuits from the perspective of brain networks is still a big challenge.
In this work, a novel decoupling generative adversarial network (DecGAN) is proposed to detect abnormal neural circuits for AD.
Concretely, a decoupling module is designed to decompose a brain network into two parts:
one part is composed of a few sparse graphs which represent the neural circuits largely determining the development of AD;
the other part is a supplement graph, whose influence on AD can be ignored.
Furthermore, the adversarial strategy is utilized to guide the decoupling module to extract the feature more related to AD.
Meanwhile, by encoding the detected neural circuits to hypergraph data, an analytic module associated with the hyperedge neurons algorithm is designed to identify the neural circuits.
More importantly,
a novel sparse capacity loss based on the spatial-spectral hypergraph similarity is developed to minimize the intrinsic topological distribution of neural circuits, which can significantly improve the accuracy and
robustness of the proposed model.
Experimental results demonstrate that the proposed model can effectively detect the abnormal neural circuits at different stages of AD,
which is helpful for pathological study and early treatment.
\end{abstract}


\section{Introduction}
\label{sec:introduction}
\IEEEPARstart{A}{lzheimer's}  disease (AD) is an irreversible and chronic neurodegenerative disease.
It is characterized by slowly progressive memory loss and cognitive deficits, constituting the most common form of dementia in the elderly and becoming a worldwide health issue\cite{a1,sh21,a2,sh6}.
According to the Alzheimer's Association\cite{AS}, it is forecasted that at least 50 million people worldwide suffer from AD, and the number will exceed 152 million by 2050 if this situation continues.
The widespread incidence of AD makes it an inevitable global issue and creates a severe financial burden to both patients and governments.
However, the causes and mechanisms of AD are still unveiled, and there aren't efficient treatments for AD.
Therefore, it is becoming increasingly important for the scientific community to develop novel methods to study the pathological features of AD.

Traditional neuroimaging methods\cite{sh2,sh8} using CT or MRI to study the morphological feature of brain regions-of-interest(ROIs) have already achieved high precision to AD's early diagnosis.
However, a nonnegligible defect of traditional imaging methods is that they can not characterize the interaction relations between ROIs.
It is the main obstacle to understand the pathological features of AD.
To overcome such barriers, new tools should be implemented.
One of the modern approaches is the analysis of the brain network.
Brain network provides a powerful representation of the interaction patterns among ROIs, which gains a lot of attention in investigating the mechanisms of AD.

A brain network can be characterized as a series of nodes and edges.
The nodes represent ROIs and edges measure regional interactions extracted from neuroimaging.
There are two common categories of brain networks: functional connectivity (FC) and structural connectivity (SC).
FC is defined as the interdependence between two ROI's blood-oxygen-level-dependent (BOLD) signals extracted from resting-state functional magnetic resonance imaging (rs-fMRI).
SC is defined as the physical connection strength between two ROI's neural fibers extracted from Diffusion Tensor Imaging (DTI).
However, most existing brain network analysis approaches focus on either FC or SC.
These approaches may ignore the complementary information existing in different modality data.
Previous studies\cite{mm1,sh9,rw1,rw2} have shown that using multimodal neuroimaging data to study brain networks can better understand the brain mechanisms.

Recent neurology studies~\cite{a10,a11} indicated that brain cognitive mechanisms involve multiple co-activated brain regions (i.e., neural circuit) interactions rather than single pairwise interactions of two ROIs.
Such neural circuits information could be critical to understanding the pathological underpinnings of AD.
However, how to detect the crucial neural circuits of AD from multimodal brain networks is still challenging:
because of the brain's many-to-one function-structure mode, traditional regression methods cannot be directly used to explore the neural circuits relationships among nodes.
Moreover, the individual variability and the non-linearity between the nodes feature and the weighted adjacent matrix of edges need to be considered simultaneously.

\begin{figure*}[!t]
\centering
\includegraphics[width=18cm]{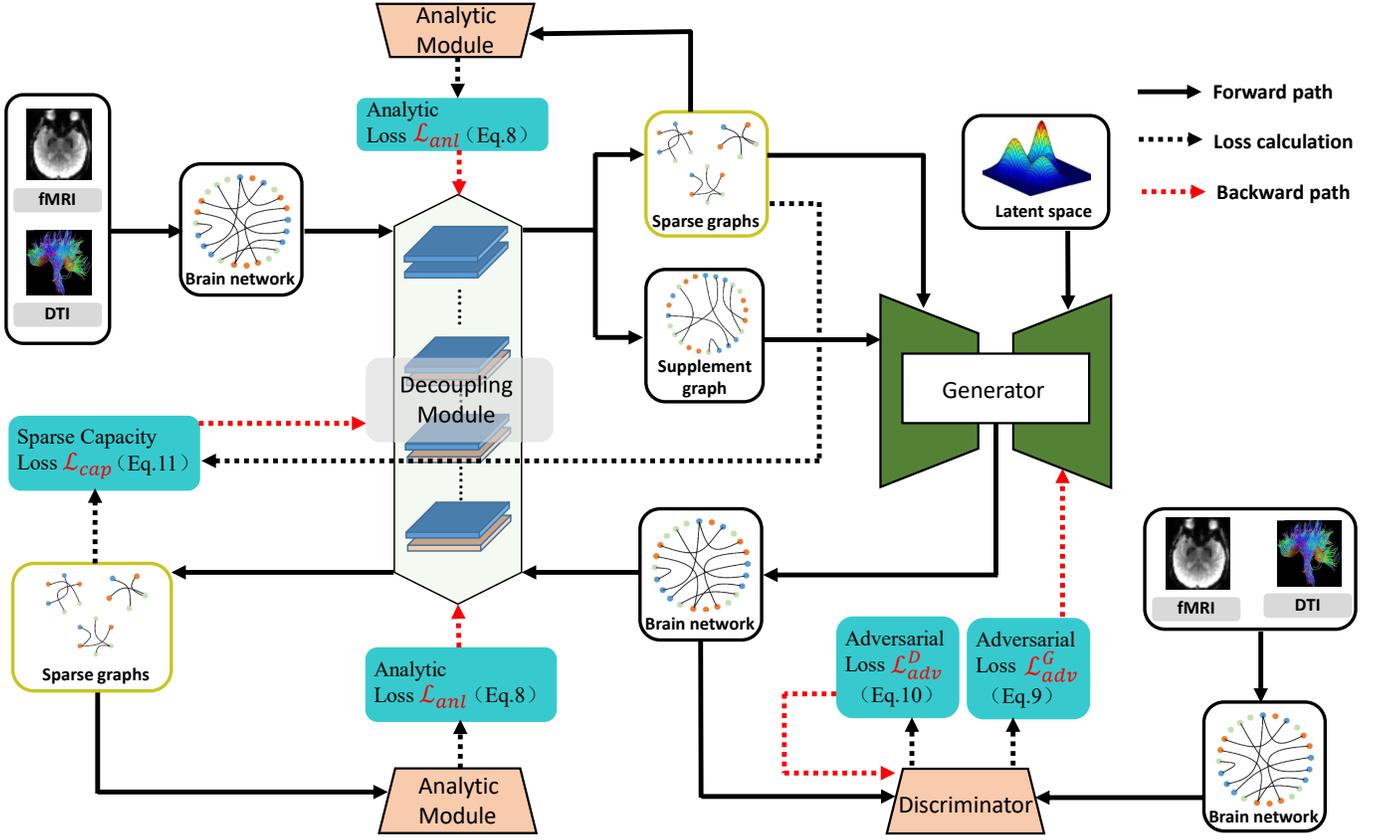}
\caption{The flowchart of DecGAN. It consists of four components: a decoupling module, a generator, a discriminator, and an analytic module.}
\label{lct}
\end{figure*}

To overcome the above-mentioned difficulties and motivated by the recent development of deep neural network based methods~\cite{sh10,sh11}, a novel decoupling
generative adversarial network (DecGAN) is proposed to detect the crucial neural circuits of AD from multimodal brain networks.
Generative adversarial networks (GANs) \cite{sh19,gan1,sh1,sh3,gan3} have attracted wide attention since it is efficient to learn complex distribution without explicitly modeling the probability density function.
It is worth mentioning that combining variational methods~\cite{sh4,sh5,sh22} with GANs can learn high-order latent boundary distribution.
This advantage of GAN can be utilized to capture the complicated high-order relationships buried in ROIs.
In this work, a novel decoupling module in DecGAN is designed to decompose the multimodal brain network into several sparse subgraphs $\{G_i\}$ and a supplement graph.
The nodes of each $G_i$ represent a crucial neural circuit of AD, and the nodes of the supplement graph represent the ROIs whose influence on AD can be ignored.
Specifically, the decoupling module is designed by using graph-based algorithms \cite{gcn1,gcn2}.
Compared to traditional CNN-based feature extracting algorithm that can only operate on regular, Euclidean data,
The graph-based algorithms can analyze interrelated and hidden structures beyond the grid neighbors, such as brain networks.
By encoding the detected neural circuits to hypergraph data,
an analytic module associated with the hyperedge neurons algorithm is designed to identify the neural circuits of the subjects, guide the decoupling module to capture the neural circuits largely determining the development of AD.
In addition, the generator of DecGAN is utilized to reconstruct the brain network by using the outputs of the decoupling module and the latent variable as inputs.
And then input the reconstruction brain network to the decoupling module, obtain sparse subgraphs $\{G_i'\}$.
Basing on hypergraph embedding and similarity of hypergraph, a novel sparse capacity loss $\mathcal{L}_{cap}$ is designed to compare the spatial-spectral differences between $\{G_i\}$ and $\{G_i'\}$ .
By minimizing $\mathcal{L}_{cap}$, the accuracy and robustness of the proposed model can be efficiently improved.
Therefore the crucial neural circuits of AD can be detected.
To the best of our knowledge, the proposed DecGAN is the first work to detect crucial neural circuits of AD by using generative adversarial networks.
The contributions of this paper are summarized as follows:
\begin{itemize}
\item[1)]
A novel decoupling module associated with an adversarial strategy is proposed to detect the abnormal neural circuits for Alzheimer's disease. By utilizing the iteration decoupling mechanism, the proposed model can efficiently extract brain network' feature highly related to Alzheimer's disease.
\end{itemize}
\begin{itemize}
\item[2)]
A novel sparse capacity loss is designed to characterize the spatial-spectral differences between two collections of neural circuits, which can minimize the difference of intrinsic topological distribution from detected neural circuits. It can significantly improve the accuracy and robustness of the proposed model.
\end{itemize}
\begin{itemize}
\item[3)]
An analytic module based on the hyperedge neurons algorithm is developed to identify the relationship between the detected neural circuits and Alzheimer's disease. The analytic module can guide the decoupling module to capture the neural circuits that largely determine the development of Alzheimer's disease.
\end{itemize}

The rest of this paper is organized as follows. The related work is reviewed in Section II. The proposed DecGAN is described in detail in Section III. In Section IV, DecGAN
is tested and compared with existing brain network analysis methods to demonstrate its advantage. Finally, concluding
remarks and future work are discussed in Section V.

\section{RELATED WORK}

With the development of machine learning (ML) techniques~\cite{sh13,sh14,sh15,sh16,sh17,sh18} in the area of brain networks,
a lot of ML models have been proposed to detect AD-related brain connectivity and predict  AD progression.
For example,
Li et al.~\cite{rw3} proposed a deep spatial-temporal feature fusion method to predict AD at its early stage.
Wang et al.~\cite{rw4} proposed a novel convolutional recurrent neural network for automated prediction of AD progression.
Zhao et al.~\cite{rw5} present a nonlinear dynamic approach to reconstruct brain networks, which shows the significant differences of FC between AD and NC.
Recently, there is a growing interest in studying brain connectivity by using graph-structured learning.
Graph-structured learning that captures the whole topology features of brain networks has significant advantages in describing the high-order relationships between ROIs.
In particular, graph convolutional networks (GCNs) is a graph-structured algorithm that generalizes convolutional neural networks (CNN)~\cite{sh12} from Euclidean data into graph-structured data~\cite{rw6,rw9,rw7}.
Recently, GCN methods have been successfully used to analyze abnormal brain connections~\cite{rw8}.
Moreover, some studies~\cite{sh7,rw10} have used spatial-based GCN to capture triplet-order relationships of brain networks.
Experiments have shown that modeling triplet-order relationships of brain networks are helpful to boost classification accuracy and learning performance.
Even though studies~\cite{rw9,rw10} just consider triplet-order relationships of brain networks,
it is a good start to study general high-order relationships of brain networks.

\section{METHODS}
\emph{Notations}: Throughout this paper, let $\mathbb{R}$ be the set of real numbers. We denote scalars, vectors,
and matrices by normal letters, lowercase boldface,
and uppercase boldface, respectively.
For a matrix $\textbf{A}$, we denote
its $(i,j)$-th entry, inverse matrix, transpose as $\textbf{A}_{i,j}$, $\textbf{A}^{-1}$, and $\textbf{A}^{T}$ ,
respectively.

\subsection{GCN and Decoupling Module}
In this study, we construct a graph to combine information on the multimodal imaging data (including rs-fMRI and DTI).
To encode the priori brain networks by rs-fMRI and DTI, let $\mathcal{G} := (\mathcal{V}, \mathcal{E})$ be an undirected weighted graph,
where $\mathcal{V}$ is the set of nodes represent 90 ROIs defined by the anatomical automatic labeling (AAL) template, $\mathcal{E}$ is the set of edges between nodes.
Let BOLD signals be the feature for the corresponding nodes, and we will denote by $\textbf{X}_{\mathcal{G}} = (X_{v_1},X_{v_2},\cdots,X_{v_{90}})$ the feature matrix .
Let the SC matrix be the weighted adjacent matrix of edges $E$, and we will denote by $\textbf{A}$ the weighted adjacent matrix.
The multi-layer GCN is defined with the following layer-wise propagation rule:
\begin{equation}\label{eqgcn}
\textbf{X}_{\mathcal{G}}^{(l+1)} = \sigma\Big(\tilde{\textbf{D}}^{-1/2}\tilde{\textbf{A}}\tilde{\textbf{D}}^{-1/2} \textbf{X}_{\mathcal{G}}^{(l)} \textbf{W}^{(l)}\Big).
\end{equation}

\begin{figure}[h!]
\centering
\includegraphics[width=0.92\columnwidth]{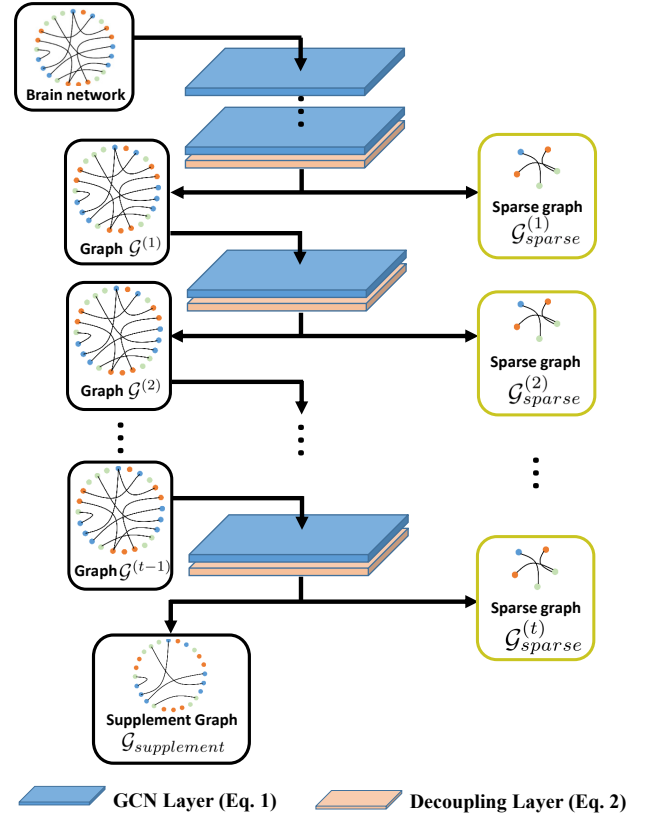}
\caption{The architecture of proposed decoupling module.}
\label{dec}
\end{figure}

Where, $\tilde{\textbf{A}} = \textbf{A} + \textbf{I}_{90}$ is the weighted adjacent matrix of $\mathcal{G}$ with added self-connections,
$\textbf{I}_{90}$ is the identity matrix of $90\times 90$, $\tilde{\textbf{D}}=\sum_{j}\tilde{\textbf{A}}_{i,j}$ is the degree matrix of $\mathcal{G}$,
and $\textbf{W}^{(l)}$ is a layer-specific trainable weight matrix.
$\sigma(\cdot)$ denotes a nonlinear activation function.
$\textbf{X}_{\mathcal{G}}^{(l)}$ is the feature matrix of nodes in the $l^{\text{th}}$ layer; $\textbf{X}_{\mathcal{G}}^{(0)}= \textbf{X}_{\mathcal{G}}$.

The decoupling layer detects a neural circuit by choosing top $k$ nodes $v_i$ which satisfy the following inequality:
\begin{equation}\label{eqdec}
\textbf{d}\cdot \sigma(\textbf{W}X_{v_i}^T+b) \geq \gamma\|\textbf{d}\|_2.
\end{equation}
Where, $\textbf{d}$ is a trainable vector, $\textbf{W}$ is a trainable weighted matrix, and $b$ is a trainable bias. $\gamma$ and $k$ are hyperparameters.
Suppose that the decoupling layer outputs a neural circuit $\mathcal{N}_1=\{v_{i_1},v_{i_2},\cdots, v_{i_m}\}$ (here $m\leq k$ since the number of nodes satisfying condition \eqref{dec} may be less than $k$),  we get the sparse graph $\mathcal{G}_{sparse}^{(1)}=(\mathcal{V},\mathcal{E}^{s(1)})$ where the weighted adjacent matrix $\textbf{A}^{s(1)}$ of edges $\mathcal{E}^{s(1)}$ is given as follows
\begin{equation}
\textbf{A}_{i,j}^{s(1)}=\left\{
\begin{aligned}
&   \textbf{A}_{i,j}, \quad \text{if}\quad (v_i,v_j)\in \mathcal{N}_1\times \mathcal{N}_1\\
&  0,  \; \; \; \; \quad \text{else}. \\
\end{aligned}
\right.
\end{equation}
Behind decoupling layer, we update $\mathcal{G}=(\mathcal{V},\mathcal{E})$ to $\mathcal{G}^{(1)}=(\mathcal{V},\mathcal{E}^{(1)})$ by setting the adjacent matrix $\textbf{A}^{(1)}=\textbf{A}-\textbf{A}^{s(1)}$.
And then using GCN layer extracts the feature matrix of $\mathcal{G}^{(1)}$, using decoupling layer gets a neural circuit $\mathcal{N}_2$ and sparse graph $\mathcal{G}_{sparse}^{(2)}$, and updates $\mathcal{G}^{(1)}$ to $\mathcal{G}^{(2)}$.
We iterate this procedure $t$-times,
we will get the neural circuits $\mathcal{N}_1,\mathcal{N}_2,\cdots,\mathcal{N}_t$. The supplementary set of $\mathcal{N}_1,\mathcal{N}_2,\cdots,\mathcal{N}_t$ will be denoted by $\mathcal{S}=\mathcal{V}\setminus (\mathcal{N}_1\cup\cdots\cup\mathcal{N}_t)$. Finally we define the supplementary graph $\mathcal{G}_{supp}=(\mathcal{V},\mathcal{E}')$ where the weighted adjacent matrix $\textbf{A}'$ of edges $\mathcal{E}'$ is given as follows
\begin{equation}
\textbf{A}'_{i,j}=\left\{
\begin{aligned}
&   \textbf{A}_{i,j}, \quad \text{if}\quad (v_i,v_j)\in \mathcal{S}\times \mathcal{S}\\
&  0,  \; \; \; \; \quad \text{else}. \\
\end{aligned}
\right.
\end{equation}
The architecture of decoupling module is shown in  Fig.\ref{dec}

\subsection{Hypergraph and Analytic Module}\label{B}
To guarantee the neural circuits that are detected by the decoupling module have a significant influence on AD,
a analytic module is designed to analyze AD progression by using these neural circuits.
The analytic module is based on a hypergraph-related algorithm.
The concepts of the hypergraph are introduced as follows.
A hypergraph is defined as $\mathcal{H}:=(\mathcal{V},\mathcal{E})$,
which includes a vertex set $\mathcal{V}$, a hyperedge set $\mathcal{E}\subset 2^{\mathcal{V}}$.
The main difference between hypergraph and graph is that a hyperedge in a hypergraph can connect more than two vertices.
A graph is a special case of a hypergraph, where each hyperedge $e$ has size $|e|=2$.
The structure of hypergraph $\mathcal{H}$ can be denoted by a $|\mathcal{V}|\times |\mathcal{E}|$ incidence matrix $\textbf{H}$, with entries defined as
\[
h(v,e)=\left\{
\begin{aligned}
&   1, \text{if}\; v\in e \\
&  0, \text{if}\; v\notin e. \\
\end{aligned}
\right.
\]
\begin{figure}[h!]
\centerline{\includegraphics[width=0.85\columnwidth]{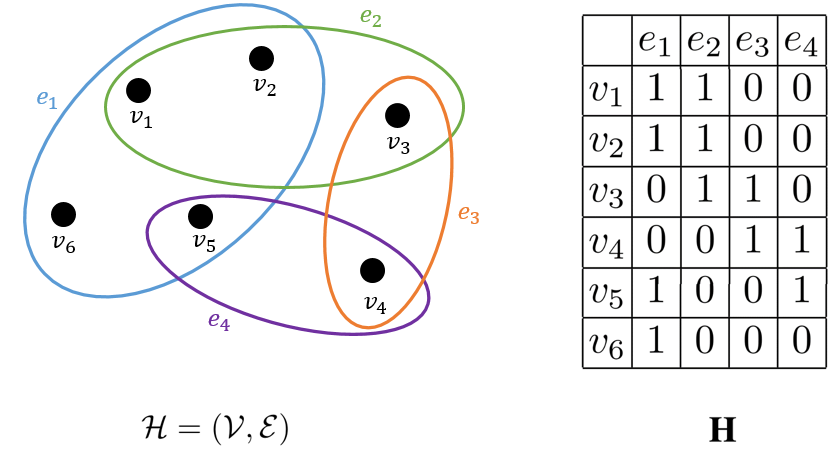}}
\caption{Left: a hypergraph $\mathcal{H}$ with $\mathcal{V}=\{v_1,v_2,v_3,v_4,v_5.v_6\}$ and $\mathcal{E}=\{e_1,e_2,e_3,e_4\}$, where $e_1=\{v_1,v_2,v_5,v_6\}$, $e_2=\{v_1,v_2,v_3\}$, $e_3=\{v_3,v_4\}$, $e_3=\{v_4,v_5\}$. Right: the incidence matrix of $\mathcal{H}$.}
\label{hyp}
\end{figure}

\begin{figure}[!t]
\centerline{\includegraphics[width=\columnwidth]{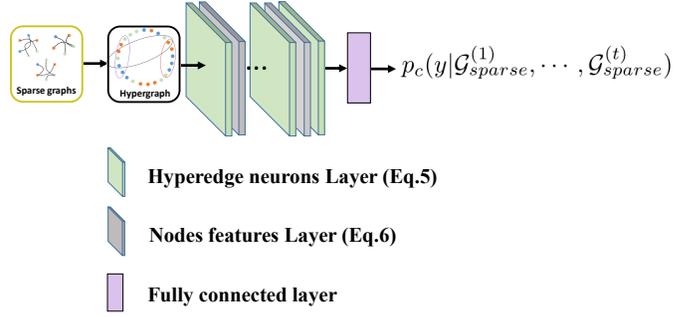}}
\caption{The architecture of proposed analytic module.}
\label{cls}
\end{figure}

Recall that each sparse graph $\mathcal{G}_{sparse}^{(p)}$ corresponds a neural circuit $\mathcal{N}_{p}$.
The collection $\{\mathcal{N}_1,\cdots,\mathcal{N}_t\}$ of neural circuits is embedded to the hypergraph $\mathcal{H}=(\mathcal{V},\mathcal{E})$ by setting $\mathcal{E}=\{\mathcal{N}_{1},\cdots,\mathcal{N}_{t}\}$.
Let the feature matrix $\textbf{X}_{\mathcal{H}}$ of the hypergraph $\mathcal{H}$ be the average of the feature matrices of the sparse graphs, i.e. $\textbf{X}_{\mathcal{H}}=\frac{1}{t}\sum_{p=1}^{t}\textbf{X}_{\mathcal{G}_{sparse}^{(p)}}$.
The hyperedge neurons algorithm proposed in~\cite{hyp1} is used to update $\textbf{X}_{\mathcal{H}}$ with the following layer-wise propagation rule:
\begin{equation}
\textbf{X}_{\mathcal{E}}^{(l)} = \sigma(\textbf{H}^{T}\textbf{X}_{\mathcal{H}}^{(l)}\textbf{W}_{\mathcal{E}}^{(l)}+b_{\mathcal{E}}^{(l)}),
\end{equation}
\begin{equation}
\textbf{X}_{\mathcal{H}}^{(l+1)} = \sigma(\textbf{H}\textbf{X}_{\mathcal{E}}^{(l)}\textbf{W}_\mathcal{V}^{(l)}+b_{\mathcal{V}}^{(l)})
\end{equation}
where $\textbf{W}_{\mathcal{E}}^{(l)}, \textbf{W}_\mathcal{V}^{(l)}$ are layer-specific trainable weight matrices, and $b_{\mathcal{E}}^{(l)}, b_{\mathcal{V}}^{(l)}$ are layer-specific trainable bias.
The architecture of the analytic module is shown in Fig.\ref{cls}.

\subsection{Generator and Discriminator}
\begin{figure}[h!]
\centerline{\includegraphics[width=\columnwidth]{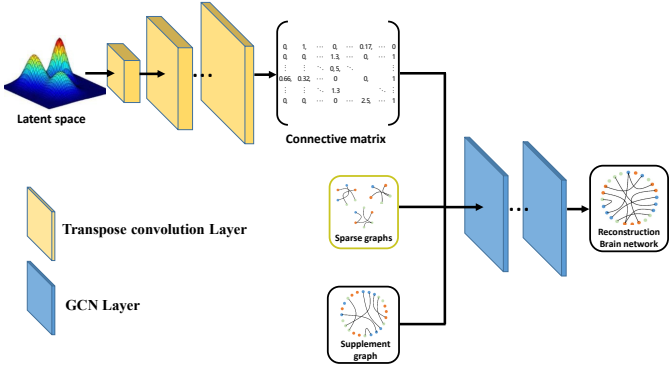}}
\caption{The network structure of proposed generator.}
\label{gd}
\end{figure}

The generator $G$ is designed to reconstruct the brain network from the sparse graphs $\mathcal{G}_{sparse}^{(1)},\cdots,\mathcal{G}_{sparse}^{(t)}$ and the supplementary graph $\mathcal{G}_{supp}$.
There are two steps in this procedure.
Firstly, we reconstruct the weighted adjacent matrix of the brain network.
Given a latent space $\mathcal{Z}$, generator learns a mapping from a random vector $z\in \mathcal{Z}$ to an output connective matrix $\overline{\textbf{A}}$.
We reconstruct the weighted adjacent matrix $\widehat{\textbf{A}}$ as follows:
\begin{equation}
\widehat{\textbf{A}}_{i,j}=\left\{
\begin{aligned}
&   \overline{\textbf{A}}_{i,j}, \quad \text{if}\quad (v_i,v_j)\notin (\cup_{p=1}^{t} \mathcal{N}_p\times \mathcal{N}_p)\cup \mathcal{S}\times \mathcal{S}\\
&  \textbf{A}_{i,j}'+\sum_{p=1}^{t}\textbf{A}_{i,j}^{s(p)},  \; \; \; \; \quad \text{else}. \\
\end{aligned}
\right.
\end{equation}
Where $\mathcal{N}_1,\mathcal{N}_2,\cdots,\mathcal{N}_t$ is the neural circuits, $\mathcal{S}$ is the supplementary set of $\mathcal{N}_1,\cdots,\mathcal{N}_t$. $\textbf{A}'$ is the weighted adjacent matrix of
supplementary graph $\mathcal{G}_{supp}$. $\textbf{A}^{s(1)},\cdots,\textbf{A}^{s(t)}$ is the weighted adjacent matrices of sparse graphs $\mathcal{G}_{sparse}^{(1)},\cdots,\mathcal{G}_{sparse}^{(t)}$, respectively.
Secondly, we use multi-layer GCN with weighted adjacent matrix $\widehat{\textbf{A}}$ and the initial feature matrix $\overline   {\textbf{X}} = \frac{1}{t+1}(\textbf{X}_{\mathcal{G}_{supp}}+\sum_{p=1}^{t}\textbf{X}_{\mathcal{G}_{sparse}^{(p)}})$ to reconstruct the feature matrix $\widehat{\textbf{X}}$ of the brain network.
The reconstructive brain network will be denoted by $G(z,\mathcal{G}_{sparse}^{(1)},\cdots,\mathcal{G}_{sparse}^{(t)},\mathcal{G}_{supp})$.
The discriminator $D$ is designed to identify whether the brain network is real or fake.
The discriminator contains of a multi-layer GCN and a fully connect layer.
The initial brain network and the reconstructive brain network are treated as real and fake samples for the discriminator training.
The structure of the generator and discriminator is illustrated in Fig.\ref{gd}

\subsection{Loss Functions}
\textbf{Analytic Loss.} Given an multimodal brain network, the goal of DecGAN is to detect the crucial neural circuits of AD that be represented by the sparse graphs.
To achieve this condition, the analytic module is introduced in~\ref{B}.
The analytic loss is imposed when optimizing the analytic module and the decoupling module.
In detail, the formula of analytic loss is defined as
\begin{equation}
\mathcal{L}_{anl} = \mathbb{E}_{\mathcal{G}}[-\log p_c(y|\mathcal{G}_{sparse}^{(1)},\cdots,\mathcal{G}_{sparse}^{(t)})].
\end{equation}
Optimize the analytic module and the decoupling module by minimizing this analytic loss.
The analytic module is trained to analyze the disease status by using sparse graphs $\{\mathcal{G}_{sparse}^{(p)}\}_{p=1}^{t}$.
The decoupling module is trained to detect the sparse graphs such that the analytic module have highest analytic accuracy when using these sparse graphs as inputs.

\textbf{Adversarial Loss.} To make the reconstructive brain network indistinguishable from initial brain net, the adversarial loss for generator $G$ is defined as
\begin{equation}
\mathcal{L}_{adv}^{G} = \mathbb{E}_{z,\mathcal{G}}[\log(D(G(z,\mathcal{G}_{sparse}^{(1)},\cdots,\mathcal{G}_{sparse}^{(t)},\mathcal{G}_{supp}))].
\end{equation}
The adversarial loss for discriminator $D$ is defined as
\begin{equation}
\begin{split}
\mathcal{L}_{adv}^{D} =& \mathbb{E}_{\mathcal{G}}[\log D(\mathcal{G})]+\\
 &\mathbb{E}_{z,\mathcal{G}}[\log(1-D(G(z,\mathcal{G}_{sparse}^{(1)},\cdots,\mathcal{G}_{sparse}^{(t)},\mathcal{G}_{supp}))].
\end{split}
\end{equation}

\textbf{Sparse Capacity Loss.} The decoupling module is optimized according to the feedback of the analytic module.
However, this procedure does not guarantees that decoupling module can detect the neural circuits with robustness when slightly changing the topological structure of brain network.
To break this dilemma, a novel sparse capacity loss $\mathcal{L}_{cap}$ is designed to characterize the structural difference of two collections of neural circuits.
Let $\{\mathcal{N}_{p}\}_{p=1}^{t}$ and $\{\mathcal{N}_{p}'\}_{p=1}^{t}$ be the collections of neural circuits which are the outputs of decoupling module using $\mathcal{G}$ and $G(z,\mathcal{G}_{sparse}^{(1)},\cdots,\mathcal{G}_{sparse}^{(t)},\mathcal{G}_{supp})$ as inputs, respectively.
The collections $\{\mathcal{N}_{p}\}_{p=1}^{t}$ and $\{\mathcal{N}_{p}'\}_{p=1}^{t}$ are embedded to the hypergraphs $\mathcal{H}=(\mathcal{V},\mathcal{E})$ and $\mathcal{H}'=(\mathcal{V},\mathcal{E}')$ by setting $\mathcal{E}=\{\mathcal{N}_{1},\cdots,\mathcal{N}_{t}\}$ and $\mathcal{E'}=\{\mathcal{N}_{1}',\cdots,\mathcal{N}_{t}'\}$, respectively.
The sparse capacity loss $\mathcal{L}_{cap}$ is consisted of two terms: The spatial similarity $\operatorname{Sim}_{spatial}(\mathcal{H},\mathcal{H}')$ and the spectral similarity $\operatorname{Sim}_{spectral}(\mathcal{H},\mathcal{H}')$ of the hypergraphs $\mathcal{H}$ and $\mathcal{H}'$, i.e.,
\begin{equation}
\mathcal{L}_{cap} = \operatorname{Sim}_{spatial}(\mathcal{H},\mathcal{H}')+\operatorname{Sim}_{spectral}(\mathcal{H},\mathcal{H}')
\end{equation}
In details, the spatial similarity is defined by
\begin{equation}
\operatorname{Sim}_{spatial}(\mathcal{H},\mathcal{H}'):=\frac{1}{t}\sum_{p=1}^{t}\frac{|\mathcal{N}_1\cap \mathcal{N}_1'|}{|\mathcal{N}_1\cup \mathcal{N}_1'|}.
\end{equation}
The spectral similarity is based on hypergraph Laplacian
\begin{equation}
\Delta_{\mathcal{H}} = \textbf{I}_{90}- \textbf{D}_{\mathcal{V}}^{-1/2}\textbf{H}\textbf{D}_{\mathcal{E}}^{-1}\textbf{H}^{T}\textbf{D}_{\mathcal{V}}^{-1/2}
\end{equation}
where $\textbf{H}$ is the incidence matrix of $\mathcal{H}$, $\textbf{D}_{\mathcal{V}}$ denotes the diagonal matrix of node degree $d(v) = \sum_{e\in \mathcal{E}}h(v,e)$,
and $\textbf{D}_{\mathcal{E}}$ denotes the diagonal matrix of edge degree $d(e) = \sum_{v\in \mathcal{V}}h(v,e)$.
Let $(\lambda_{\mathcal{H}}^{(1)},\cdots,\lambda_{\mathcal{H}}^{(90)})$ and $(\lambda_{\mathcal{H}'}^{(1)},\cdots,\lambda_{\mathcal{H}'}^{(90)})$ be the eigenvalues of the hypergraph Laplacian $\Delta_{\mathcal{H}}$ and $\Delta_{\mathcal{H'}}$, respectively.
The spectral similarity of $\mathcal{H}=(\mathcal{V},\mathcal{E})$ and $\mathcal{H}'=(\mathcal{V},\mathcal{E}')$ is defined by
\begin{equation}
\operatorname{Sim}_{spectral}(\mathcal{H},\mathcal{H}'): =\frac{1}{90}\sum_{i=1}^{90}|\lambda_{\mathcal{H}}^{(i)}-\lambda_{\mathcal{H}'}^{(i)}|^2.
\end{equation}

\textbf{Total Loss}. The total loss functions to optimize the decoupling module $M$, the analytic module $A$, the generator $G$, and the discriminator $D$ are summarized respectively as
\begin{equation}
\mathcal{L}_{M} = \mathcal{L}_{anl} + \gamma\mathcal{L}_{cap}
\end{equation}
\begin{equation}
\mathcal{L}_{A} = \mathcal{L}_{anl}
\end{equation}
\begin{equation}
\mathcal{L}_{G} = \mathcal{L}_{adv}^{G}
\end{equation}
\begin{equation}
\mathcal{L}_{D} = \mathcal{L}_{adv}^{D}
\end{equation}
where $\gamma$ is hyperparameter that control the relative importance of analytic loss and sparse capacity loss.

\section{EXPERIMENTS}
\subsection{Dataset and Preprocessing}
DTI and rs-fMRI images from the Alzheimer's Disease Neuroimaging Initiative (ADNI) public dataset are used to validate our proposed framework.
There are 236 subjects' data used in our study. The detailed subjects' information is summarized in Table~\ref{information}.

\begin{table}[h]
\caption{Subjects information in this study}

\label{information}

\centering
\begin{tabular}{c c c c c c}

\toprule[2pt]

  \textbf{Group} & \textbf{AD}(53) & \textbf{LMCI}(31) & \textbf{EMCI}(74) & \textbf{NC}(78) \\

\midrule[1pt]

  Male/Female & 32M/21F & 15M/16F & 45M/29F & 35M/43F   \\

  Age(mean $\pm$ SD)  & 75.3 $\pm$ 5.5 & 74.9 $\pm$ 5.3 & 75.8 $\pm$ 6.1 & 74.0 $\pm$ 5.9   \\

\bottomrule[2pt]

\end{tabular}

\end{table}

In this study, DPARSF toolbox~\cite{dparsf} and GRETNA~\cite{gretna} toolbox is used to preprocess the rs-fMRI data.
At first, the initial DICOM format of rs-fMRI data is converted to NIFTI format data.
Then, standard steps for rs-fMRI data preprocessing is applied by using the DPARSF, including the discarding of the first 20 volumes, head
motion correction, spatial normalization, and Gaussian smoothing in this stage.
Next, the AAL atlas is used to divide brain space into 90 ROIs.
Finally, the time series of all voxels are extracted by using the GRETNA.
By taking the mean value of the time series of all voxels in a specific ROI, the BOLD signal of individual ROI is obtained.

PANDA toolbox~\cite{panda} is used to preprocess the DTI data.
Similar to the preprocessing of rs-fMRI data, at first, the initial DICOM format of DTI data is converted to NIFTI format data.
Then, skull stripping, resampling the fiber bundle, head motion correction is applied.
Next, the AAL atlas is used to divide brain space into 90 ROIs.
Each ROI is defined as a node of the brain network.
Finally, the structural connectivity of the brain network is determined by fiber tracking between different ROIs.
The fiber tracking stopping condition is defined as follows:
(1) the crossing angle between two consecutive moving directions is more than 45 degrees.
(2) the fractional anisotropy value is not in the range of [0.2, 1.0].

\subsection{Experiment Settings}
A five-fold validation strategy is used to evaluate the performance of our proposed DecGAN model.
In detail, all subjects are randomly divided into five subsets with equal size.
One of the subsets is treated as the test set, and the union of the other four subsets is treated as the training set.
Repeat this process five times to remove the bias by the random division.
The classification performance is evaluated by mean values of detection accuracy(ACC), sensitivity(SEN), specificity(SPEC), and F-score(F1).
The proposed DecGAN is implemented in Pytorch.
All experiments in this study are conducted on four NVIDIA GeForce GTX 2080 Ti GPUs.
The optimizer is set to 'Adam'.
The batch size is set to 16.
The decoupling coefficient $\gamma$ is set to 0.1.
The learning rate of the decoupling module, analytic module, generator, and discriminator are set to $10^{-4}$, $10^{-3}$, $10^{-4}$, and $10^{-4}$, respectively.

\subsection{Neural Circuits Analysis}
\begin{figure}[h!]
\centerline{\includegraphics[width=\columnwidth]{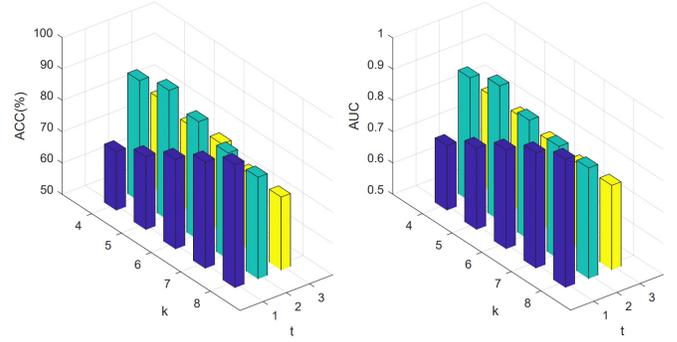}}
\caption{Left: ACC comparison with various of $t$ and $k$; Right: AUC comparison with various of $t$ and $k$}
\label{acc}
\end{figure}

\begin{figure}[h!]
\centerline{\includegraphics[width=\columnwidth]{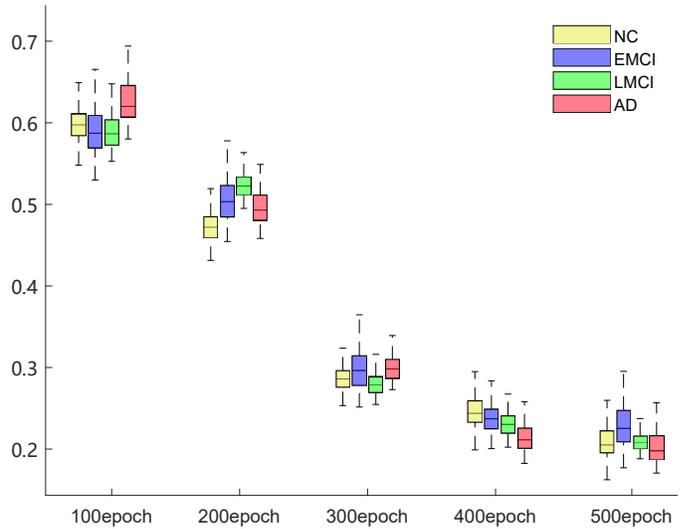}}
\caption{Root Mean-Square Error between the reconstruction structural connectivity and the priori structural connectivity}
\label{box}
\end{figure}

In this section, the experiments are conducted to study the influence of the hyper-parameters $t$ and $k$ of the decoupling module on classification performance.
Recall that the hyper-parameter $t$ represents the number of neural circuits which are outputted by the decoupling module, and the hyper-parameter $k$ represents the maximal number of ROIs contained in each neural circuit.
The experiments evaluate the ACC and AUC value of binary classification tasks (AD vs. NC) with varied parameters $t = [1,2,3]$, $k = [4,5,6,7,8]$ while keeping other parameters invariant.
The results are shown in Fig.~\ref{acc}.
The top two highest ACC value of $92.59\%$ and $88.90\%$ are achieved at $(t,k)=(2,5)$ and $(t,k)=(1,8)$.
In addition, root mean square error (RMSE) is used to measure the structural differences between the priori brain network and the reconstruction brain network.
Thus, to generate reliable reconstruction brain network, the RMSE should keep decreasing before converged.
The quantitative analysis between the reconstruction brain networks and the priori brain networks is shown in Fig.\ref{box}.

\begin{figure*}[htb]
\centering
\includegraphics[width=18cm]{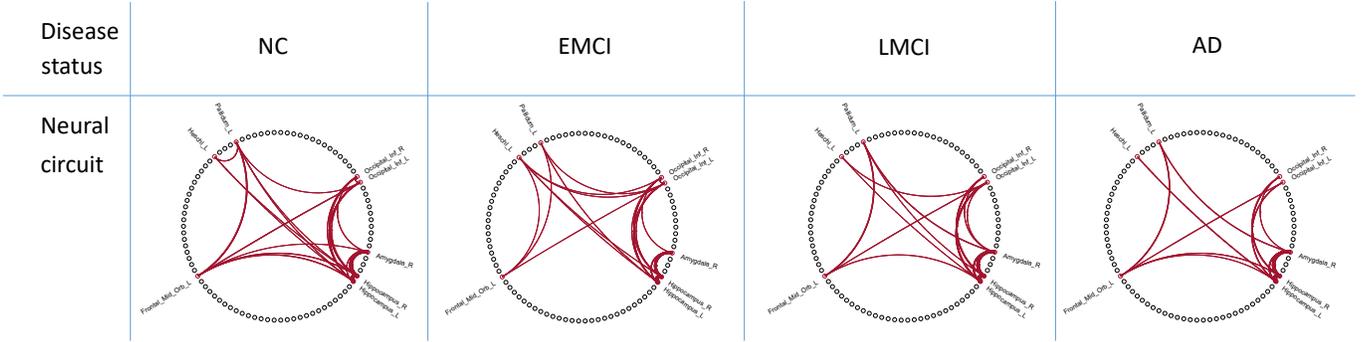}
\caption{The connectivity of the neural circuit with the hyperparameters $(t,k)=(1,8)$.}
\label{c8}
\end{figure*}

\begin{figure*}[t]
\centering
\includegraphics[width=18cm]{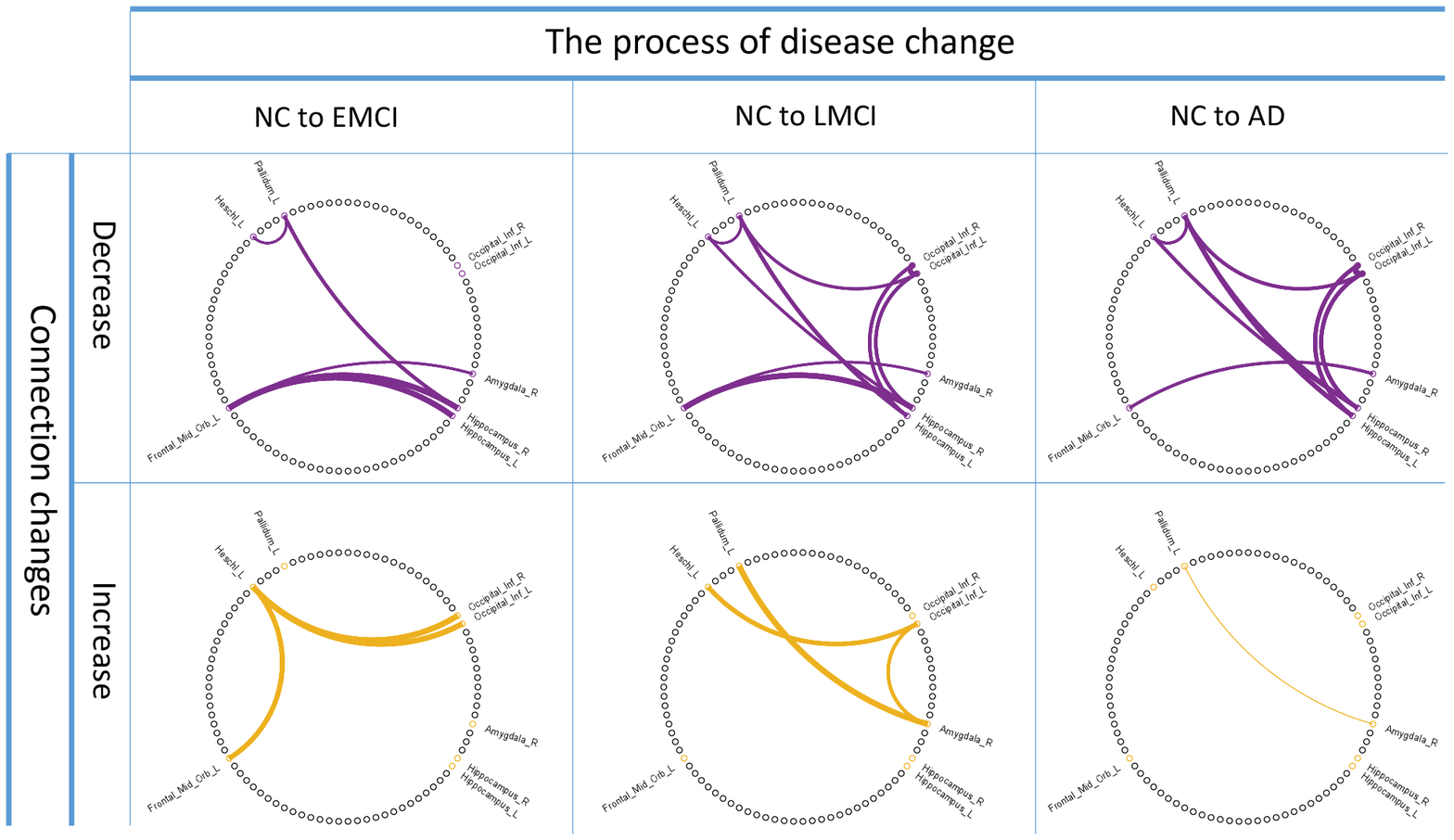}
\caption{The connectivity changes of the neural circuit between two disease statuses with the hyperparameters $(t,k)=(1,8)$. }
\label{v1}
\end{figure*}

\begin{figure*}[t]
\centering
\includegraphics[width=18cm]{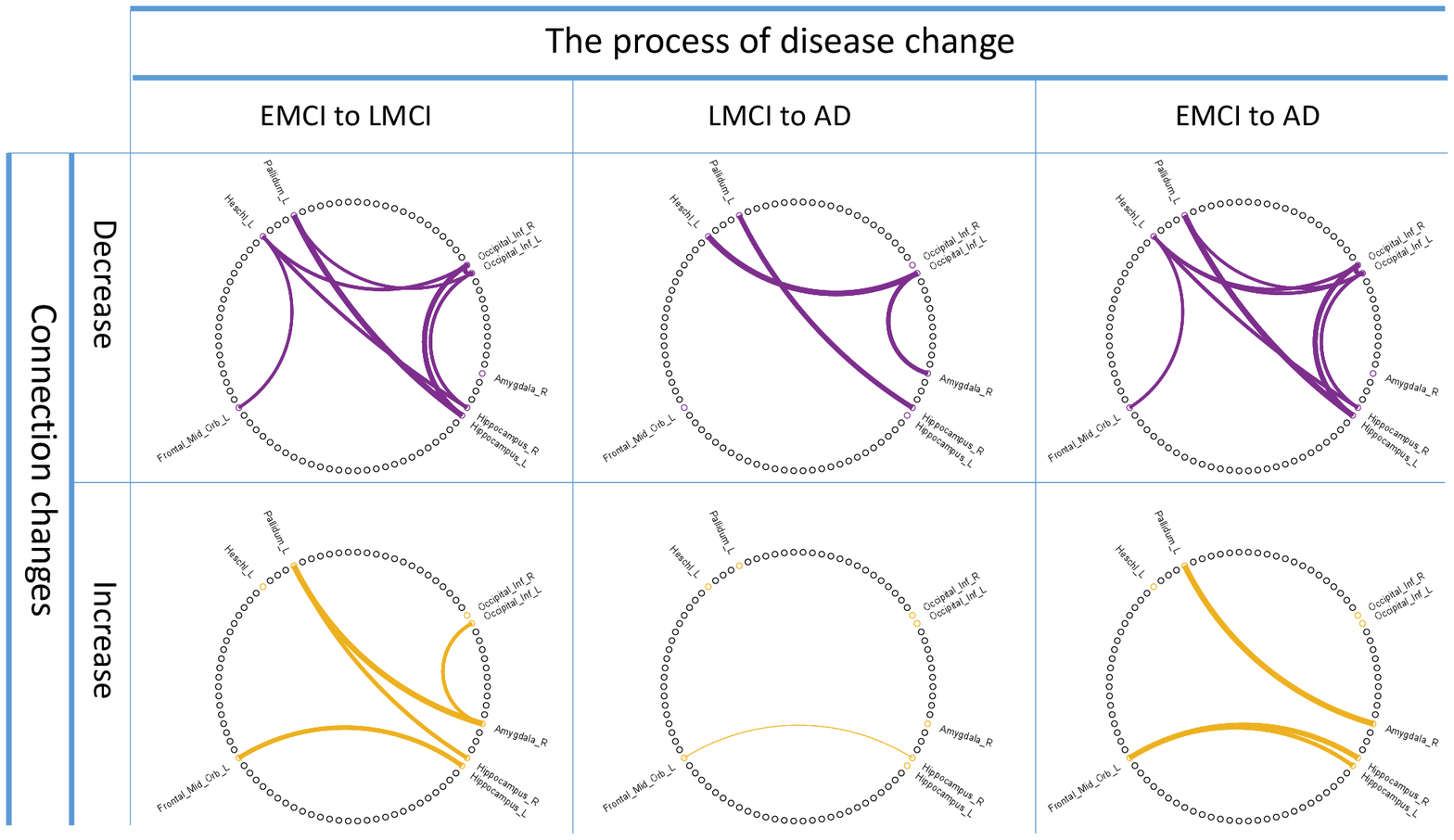}
\caption{The connectivity changes of the neural circuit between two disease statuses with the hyperparameters $(t,k)=(1,8)$.}
\label{v2}
\end{figure*}

\begin{figure*}[t]
\centering
\includegraphics[width=18cm]{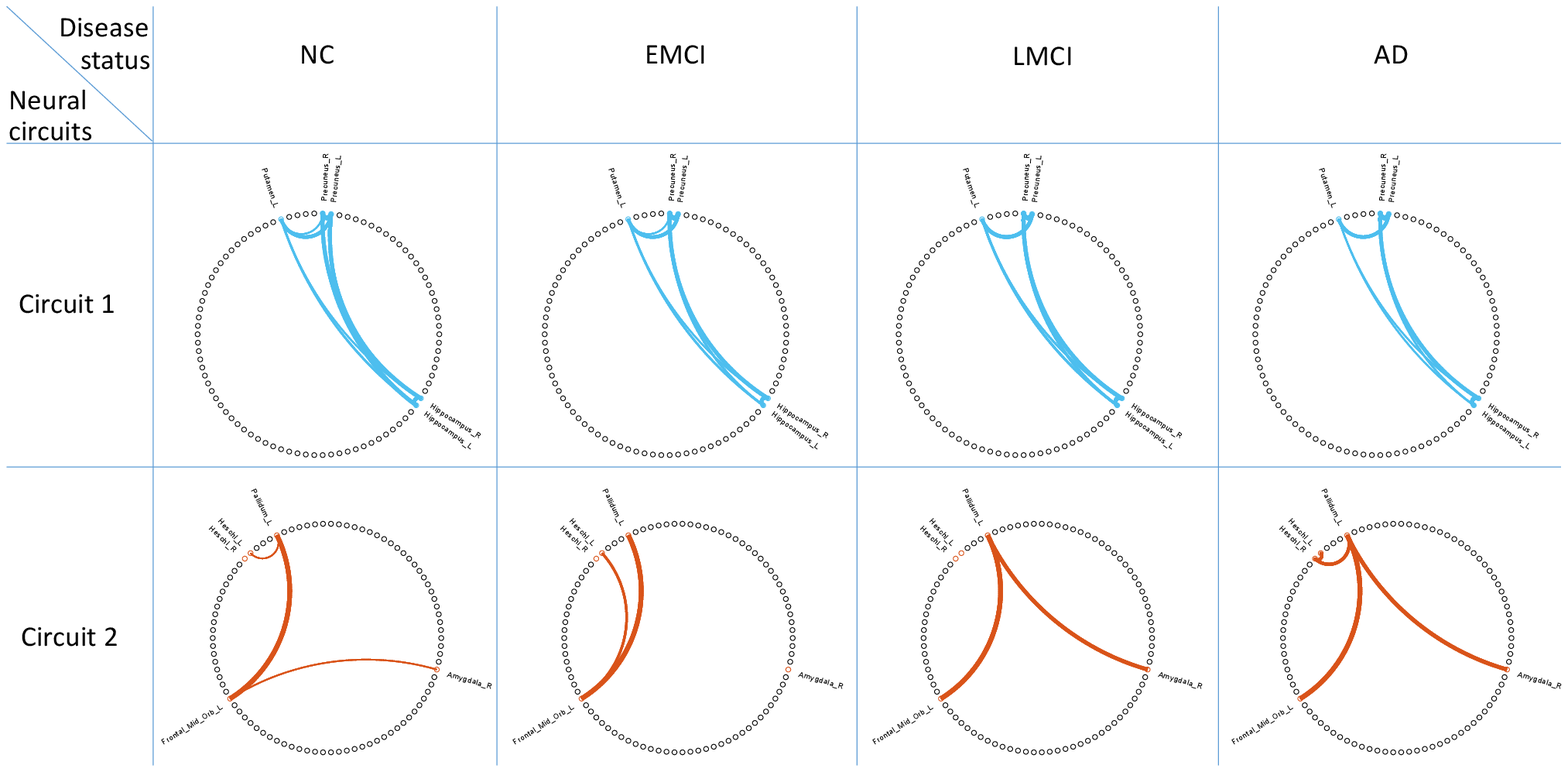}
\caption{The connectivity of the neural circuits with the hyperparameters $(t,k)=(2,5)$.}
\label{c5}
\end{figure*}

\begin{figure*}[t]
\centering
\includegraphics[width=18cm]{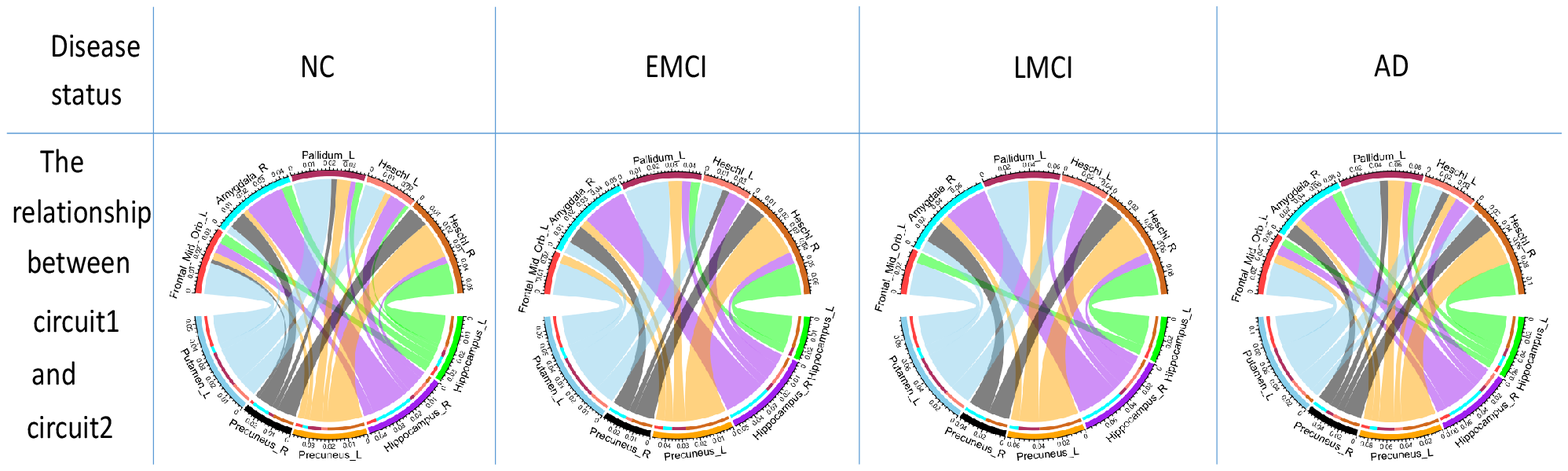}
\caption{This figure shows the interaction of two neural circuits with the hyperparameters $(t,k)=(2,5)$. The lower semicircle and upper semicircle represents the neural circuit 1 and the neural circuit 2 shown in Fig.\ref{c5}, respectively. the width of bands represents the connection strength between ROI in the lower semicircle and ROI in the upper semicircle.}
\label{r}
\end{figure*}

Counting all the output from total subjects, the abnormal neural circuits for AD is $\big\{$[Hippocampus\_L, Hippocampus\_R, Precuneus\_L, Precuneus\_R, Putamen\_L];[Frontal\_Mid\_Orb\_L, Amygdala\_R, Pallidum\_L, Heschl\_L, Heschl\_R]$\big\}$ when $(t,k)=(2,5)$.
The abnormal neural circuits for AD is $\big\{$[Frontal\_Mid\_Orb\_L, Hippocampus\_L, Hippocampus\_R, Amygdala\_R, Occipital\_Inf\_L, Occipital\_Inf\_R, Pallidum\_L, Heschl\_L]$\big\}$ when $(t,k)=(1,8)$.
The structural connectivity in neural circuits are shown in Fig.\ref{c8},\ref{v1},\ref{v2},\ref{c5},\ref{r}.

\subsection{Ablation analysis}
We propose the sparse capacity loss to compare the structural differences between the neural circuits decoupled from the priori brain network and the neural circuits decoupled from the reconstruction
brain network. In order to explore the effectiveness of the sparse capacity loss, two ablation experiments are performed in this paper.
\begin{figure}[h!]
\centerline{\includegraphics[width=0.85\columnwidth]{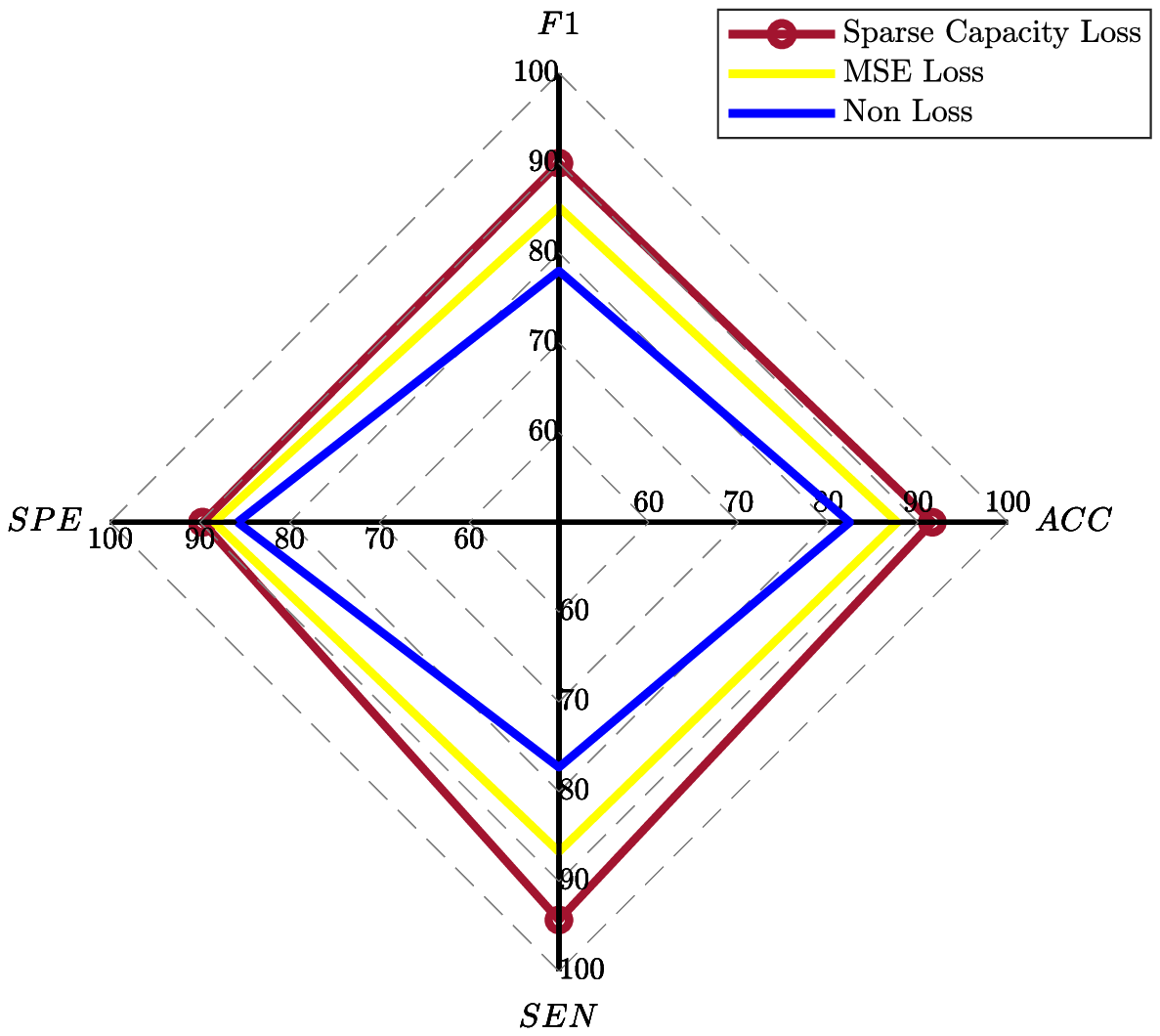}}
\caption{The comparison of classification performance of AD vs. NC with different loss functions.}
\label{radar}
\end{figure}
One is using MSE loss to replace the sparse capacity loss to measure the structural difference between two neural circuits.
The other is directly removing the sparse capacity loss.

The effects of the sparse capacity loss on classification performance is shown in Fig.\ref{radar}.
As shown in Fig.\ref{radar}, the results of the sparse capacity loss are better than MSE loss and without loss function.
The improvement of ACC, SEN, SPE, and F1 value demonstrate that the sparse capacity loss can effectively enhance the performance of the decoupling module.


\section{Discussion and conclusion}
\subsection{Comparison with clinical results}
By counting the number of occurrences of ROIs in the neural circuits of model output, we get the top ten ROIs, specificity, these ROIs are Frontal\_Mid\_Orb\_L, Hippocampus\_L, Hippocampus\_R, Amygdala\_L, Occipital\_Inf\_L,
Occipital\_Inf\_R, Precuneus\_L, Precuneus\_R, Heschl\_L and Heschl\_R.
We can see that these brain regions are mainly concentrated on the memory and reasoning areas, which are highly related to the AD according to the clinical studies~\cite{cl1,sh20}.
\begin{table}[h!]\normalsize
\caption{The clinical information of the top ten AD-related ROIs detected by the proposed model.}
\setlength{\tabcolsep}{3pt}

\begin{tabular}{|p{82pt}|p{68pt}|p{82pt}|}
\hline
\rowcolor{black}
\textcolor[rgb]{1.00,1.00,1.00}{\textbf{AAL region}}&
\textcolor[rgb]{1.00,1.00,1.00}{\textbf{Location}}&
\textcolor[rgb]{1.00,1.00,1.00}{\textbf{Evidence}}  \\
\hline
\rowcolor{mygray}

Frontal\_Mid\_Orb\_L&
Frontal lobe&
Salat \emph{et al}. \cite{roi9} \\
\hline

Hippocampus\_L&
Limbic lobe&
Du \emph{et al}. \cite{roi37} \\
\hline
\rowcolor{mygray}

Hippocampus\_R&
Limbic lobe&
Du \emph{et al}. \cite{roi37} \\
\hline

Amygdala\_L&
Limbic lobe&
Tsuchiya \emph{et al}. \cite{roi42} \\
\hline
\rowcolor{mygray}

Occipital\_Inf\_L&
Occipital lobe&
Sun \emph{et al}. \cite{roi53} \\
\hline

Occipital\_Inf\_R&
Occipital lobe&
Gupta \emph{et al}. \cite{roi54} \\
\hline
\rowcolor{mygray}

Precuneus\_L&
Parietal lobe&
Yang \emph{et al}. \cite{roi67} \\
\hline

Precuneus\_R&
Parietal lobe&
Karas \emph{et al}. \cite{roi68} \\
\hline
\rowcolor{mygray}

Heschl\_L&
Temporal lobe&
Zhou \emph{et al}. \cite{roi79} \\
\hline

Heschl\_R&
Temporal lobe&
Pusil \emph{et al}. \cite{roi80} \\
\hline
\end{tabular}

\label{evidence}
\end{table}

\begin{table*}[t!]
\caption{Prediction performance of the priori brain networks and reconstruction brain networks under different classifiers.}\label{dissclf}
\resizebox{\textwidth}{15.5mm}{
\begin{tabular}{|l|l|l|l|l|l|l|l|l|l|l|l|l|l|}
	\hline
	\textbf{Classifier} \quad & \textbf{Input}  \quad & \multicolumn{4}{c|}{\textbf{NC vs. EMCI}}  & \multicolumn{4}{c|}{\textbf{NC vs. LMCI}} & \multicolumn{4}{c|}{\textbf{NC vs. AD}} \\ \cline{3-14}
    \quad & \quad & ACC \; & SEN \; & SPE \; & F1 \;& ACC \;& SEN \;& SPE \; & F1 \;& ACC \;& SEN \;& SPE \; & F1 \;\\
    \hline
    \hline
    SVM & Priori brain networks & 72.41\% & 73.33\% & 71.42\% & 73.33\% & 76.19\% & 86.66\% &  50.00\% & 83.87\% & 77.77\% & \textbf{87.50\%} & 63.63\% & 82.35\% \\ \cline{2-14}
     & Reconstruction brain networks & \textbf{75.86\%} & \textbf{80.00\%} & 71.42\% & \textbf{77.41\%} & \textbf{80.95\%} & 86.66\% &  \textbf{66.66\%} & \textbf{86.66\%} & \textbf{85.18\%} & 75.00\% & \textbf{100.00\%} & \textbf{85.71\%} \\ \cline{2-14}
    \hline
    \hline
    DNN & Priori brain networks & 79.31\% & \textbf{86.66\%} & 71.42\% & \textbf{81.25\%} & 80.95\% & 80.00\% &  83.33\% & 85.71\% & 81.48\% & 87.50\% & 72.72\% & 84.85\%  \\ \cline{2-14}
    & Reconstruction brain networks & 79.31\% & 73.33\% & \textbf{85.71\%} & 78.57\% & \textbf{85.71\%} & \textbf{86.66\%} &  83.33\% & \textbf{89.65\%} & \textbf{85.18\%} & \textbf{93.75\%} & 72.72\% & \textbf{88.23\%} \\ \cline{2-14}
    \hline
    \hline
    GCN & Priori brain networks & 82.75\% & 86.66\% & 78.57\% & 83.87\% & 80.95\% & 73.33\% &  100.00\% & 84.61\% & 81.48\% & 81.25\% & 81.80\% & 83.87\% \\ \cline{2-14}
     & Reconstruction brain networks & \textbf{86.20\%} & 86.66\% & \textbf{85.71\%} & \textbf{86.66\%} & \textbf{85.71\%} & \textbf{80.00\%} & 100.00\% & \textbf{88.88\%} & \textbf{85.18\%} & 81.25\% & \textbf{90.91\%} & \textbf{86.66\%} \\ \cline{2-14}
    \hline
\end{tabular}}
\end{table*}

In detail, literature verification is carried out to find out whether these ROIs are related to AD,
the results are shown in Table~\ref{evidence}.
Note that these ROIs are mainly located in the Limbic lobe, Occipital lobe, Parietal lobe, and Temporal lobe.
The limbic lobe is regarded as an important brain area that highly relates to AD pathology~\cite{limbic}.
The occipital lobe is significantly related to visual memory,  in which AD patients showed atrophy in the occipital cortex~\cite{occipital}.
The parietal lobe is involved in the spatial function and is particularly important for real-time spatial navigation, in which AD patients showed the change in parietal lobe white matter hyperintensities~\cite{parietal}.
The parietal lobe plays important role in integrating sensory information from various parts of the body, knowledge of numbers. The AD patients showed rapid atrophy in the medial temporal lobe~\cite{temporal}.
Therefore, these clinical results prove the High correlation of the neural circuits detected by our model to AD.

\subsection{The Reconstruct Brain Networks By Generator}
\begin{figure}[h!]
\centerline{\includegraphics[width=\columnwidth]{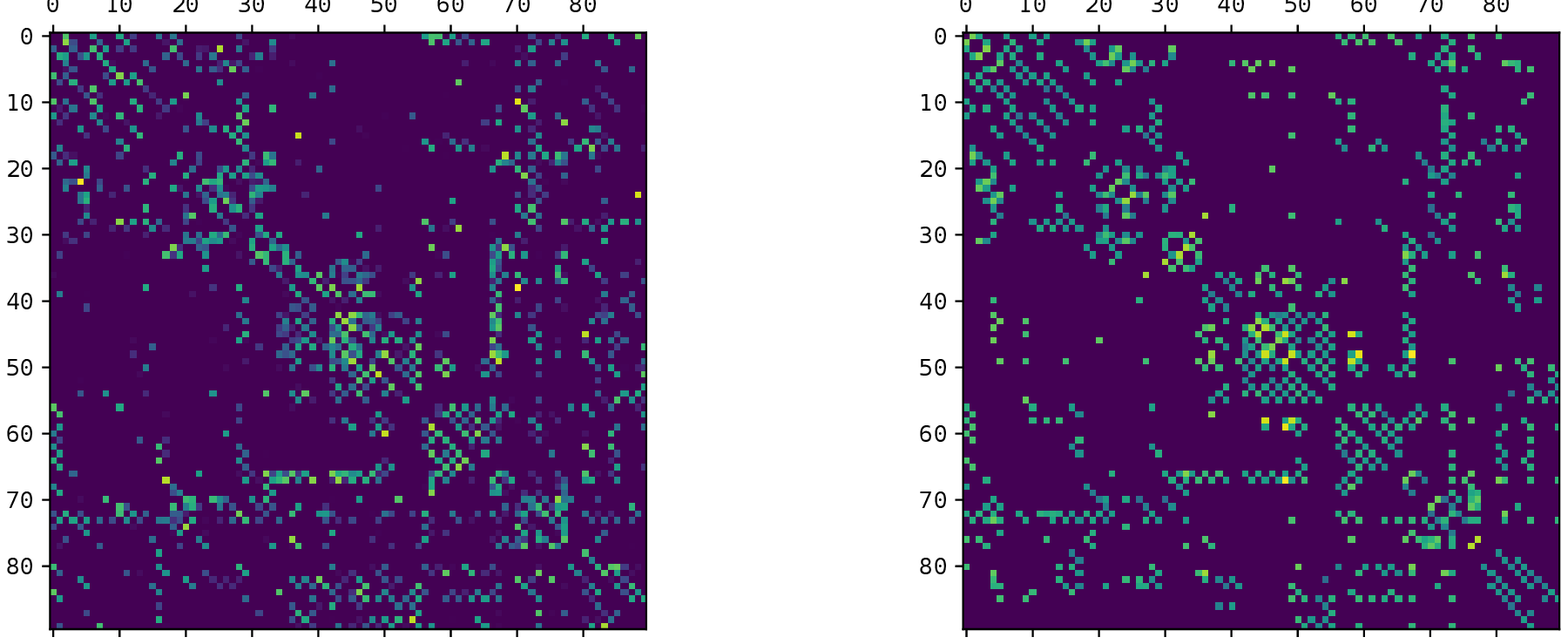}}
\caption{Left: the reconstruction structural connectivity; Right: the priori structural connectivity}
\label{generator}
\end{figure}

The generator is used to reconstruct the brain network from the output of the decoupling module and the latent space.
The main goal of the proposed model is to detect the abnormal neural circuits which have a significant influence on AD.
The stability of the decoupling module can be improved by enhancing the AD-related feature expression ability of the reconstruction brain networks.
Therefore, the quality of the reconstruction brain network is the key factor that affects whether the proposed model can detect neural circuits accurately.
To verify the AD-related feature expression ability of the reconstruction brain networks, the classification experiment is designed to compare the prediction performance of the priori brain networks and reconstruction brain networks on different AD stages.
The classification results is shown in Table~\ref{dissclf}.
Finally, the visualization of the structural connectivity (i.e. weighted adjacent matrix of brain network) of the brain networks is shown in Fig.\ref{generator}.

\subsection{Conclusion}
In this paper, we propose a novel decoupling generative adversarial network (DecGAN) to detect the crucial neural circuits for AD.
Benefit from the GCN layer and the decoupling layer, the proposed model can efficiently extract complementary topology information between rs-fMRI and DTI.
Moreover, we propose the sparse capacity loss to characterize the intrinsic topological difference between different neural circuits, which significantly improved the robustness and accuracy of the proposed model.
This paper only focuses on AD, but it is worth mention that the proposed model can be easily extended to other neurodegenerative diseases.
Although the proposed model is promising in providing a new multimodal analysis framework for neural circuits detection,
there are still two major limitations in our works. One limitation is that the proposed model cannot explain how the internal mechanisms of the detected neural circuits affect the development of the disease.
A possible solution is introducing recurrent neural networks to analyze the BOLD signal, which makes it possible to quantitatively characterize the influence of neural circuits on the disease process.
Another limitation is that our current study dataset is relatively small.
In the future, we plan to test the effectiveness of the proposed model on a larger dataset of brain images such as UKBiobank.


\end{document}